\documentclass{article}

\usepackage[preprint]{neurips_data_2024}

\usepackage[utf8]{inputenc} 
\usepackage[T1]{fontenc}    
\usepackage{hyperref}       
\usepackage{url}            
\usepackage{booktabs}       
\usepackage{amsfonts}       
\usepackage{nicefrac}       
\usepackage{microtype}      
\usepackage{xcolor}         
\usepackage{pifont}
\usepackage{multicol}
\usepackage{multirow}
\usepackage{colortbl}
\usepackage{array}
\usepackage{graphicx}
\usepackage{makecell}
\usepackage{caption}
\usepackage{subcaption}
\usepackage{longtable}
\usepackage{booktabs}

\usepackage{graphicx}
\usepackage{float} 
\usepackage{subcaption}
\usepackage{amsmath}

\title{SurvMamba: State Space Model with Multi-grained Multi-modal Interaction for Survival Prediction}

\author{%
Ying Chen\textsuperscript{1}\thanks{Equal contribution.}
\qquad Jiajing Xie\textsuperscript{2}\footnotemark[1]
\qquad Yuxiang Lin\textsuperscript{2}
\qquad Yuhang Song\textsuperscript{1}
\\
\textbf{Wenxian Yang\textsuperscript{3}}
\qquad \textbf{Rongshan Yu\textsuperscript{1,2\thanks{Corresponding author.}}}\\
\\
\textsuperscript{1}School of Informatics, Xiamen University \quad \hspace{0.5em}  
\\
\textsuperscript{2}National Institute for Data Science in Health and Medicine, Xiamen University\\ 
\textsuperscript{3}Aginome Scientific \\
{\tt\small rsyu@xmu.edu.cn}
}

\begin{document}

\maketitle

\vspace{1cm} 

\begin{abstract}
Multi-modal learning that combines pathological images with genomic data has significantly enhanced the accuracy of survival prediction. 
Nevertheless, existing methods have not fully utilized the inherent hierarchical structure within both whole slide images (WSIs) and transcriptomic data, from which better intra-modal representations and inter-modal integration could be derived. 
Moreover, many existing studies attempt to improve multi-modal representations through attention mechanisms, which inevitably lead to high complexity when processing high-dimensional WSIs and transcriptomic data. Recently, a structured state space model named Mamba emerged as a promising approach for its superior performance in modeling long sequences with low complexity. In this study, we propose Mamba with multi-grained multi-modal interaction (\textbf{SurvMamba}) for survival prediction. 
SurvMamba is implemented with a Hierarchical Interaction Mamba (HIM) module that facilitates efficient intra-modal interactions at different granularities, thereby capturing more detailed local features as well as rich global representations. 
In addition, an Interaction Fusion Mamba (IFM) module is used for cascaded inter-modal interactive fusion, yielding more comprehensive features for survival prediction. Comprehensive evaluations on five TCGA datasets demonstrate that SurvMamba outperforms other existing methods in terms of performance and computational cost. 
\end{abstract}

\section{Introduction}

Survival prediction evaluates patients' mortality risks, thereby enhancing the clinical decision-making process related to diagnosis and treatment planning~\cite{tran2021deep}.
For cancer patients, pathological images and genomic profiles provide critical and interconnected information for patient stratification and survival analysis~\cite{shao2019integrative}. For example, pathological images detail the tumor microenvironment, capturing the diversity of cancer cells and immune interactions~\cite{gui2023multimodal}. At the same time, genomic profiles provide critical insights into cancer cell states and immune system factors that affect dynamic prognostic outcomes~\cite{milanez2020cancer, tong2023prioritizing}. Therefore, integrating pathological images and genomic data through multi-modal learning holds considerable promise to enhance the precision of cancer survival predictions.

Extensive efforts have been made in multi-modal survival analysis with histological whole slide images (WSIs) and transcriptomic data~\cite{jaume2023modeling, xu2023multimodal, subramanian2021multimodal}. Among them, Multiple Instance Learning (MIL)~\cite{jaume2023modeling, xu2023multimodal, qiu2023deep} has emerged as an effective method to process the high-dimensional WSIs and transcriptomic data. In these MIL-based methods, each WSI is represented as a ``bag'' with numerous patches as instances, while transcriptomic data is organized into a ``bag'' with functional genomes~(\emph{i.e.}, genomic groups) as instances. Subsequently, fusion of histological and genomic features~\cite{li2022hfbsurv, chen2022pan} is conducted to predict survival outcomes. 

\begin{figure*}[t]
  \centering
  \includegraphics[width=0.8\linewidth]{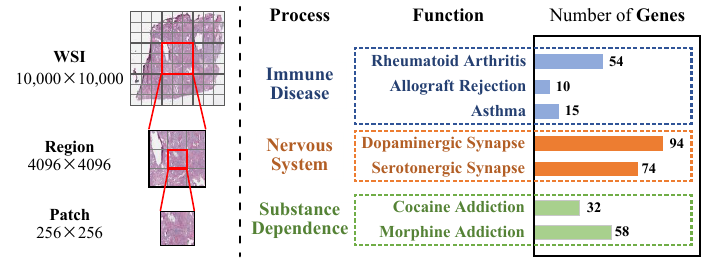}
  \caption{Hierarchical structure of WSI and transcriptomic data. }
  \label{fig:Hierarchical}
\end{figure*}

Despite the complex, high-dimensional nature of WSIs and transcriptomic data, they exhibit significant inherent hierarchical structures. These structures stem from their fundamental biological functions and the pathology associated with diseases, as illustrated in Figure~\ref{fig:Hierarchical}. WSI reveals tissue organization at region-level and provides detailed cellular insights at patch-level. Genes can be categorized according to their genomic function, which can be further subdivided based on biological process. However, existing methods do not fully leverage these hierarchical structures in both WSIs and transcriptomic data~\cite{chen2022scaling, kanehisa2000kegg}, which may lead to the following issues. 
The first issue regards to insufficient global representation. There are multi-level prognostic insights reflected in the hierarchical information. For example, fine-grained patch-level pathological images detail cell densities~\cite{zhang2024inferring}, while coarse-grained region-level images reveal tissue features and tumor-immune interactions~\cite{fu2020pan}. Fine-grained function-level transcriptomics reveals the specific functionalities of gene sets~\cite{ietswaart2021genewalk}, while coarse-grained process-level data focus on identifying those fundamental macroscopic biological processes~\cite{zhang2021protocol}. Therefore, important global features for survival prediction could potentially be lost if a method focuses on fine-grained information only. 
Furthermore, there is limited cross-modality communications. Both WSIs and transcriptomic data contain a multitude of cross-modality communications at different hierarchical levels, including 
patch-to-function and region-to-process relationships~\cite{ke2023mine, smug2023ongoing}, which cannot be easily revealed if cross-modality communications are only established at fine-grained features level, leading to deficient inter-modal integration.

Although the integration of hierarchical structures into multi-modal survival prediction frameworks holds promise for yielding more comprehensive prognostic insights, it poses significant computational challenges, especially when implemented using attention mechanisms, given their high computational complexity. On the other hand, the Selective Structured State Space Model (SSM), known as Mamba~\cite{gu2023mamba}, has demonstrated remarkable efficiency in long sequence modeling. Mamba effectively captures long-range dependencies and improves training and inference efficiency through a selection mechanism and a hardware-aware algorithm, hence providing an alternative to deal with high-dimensional data.

In this paper, we propose a Mamba-based survival prediction method with multi-grained multi-modal interaction, termed \textbf{SurvMamba}, which extracts multi-modal multi-grained information from hierarchical structure and facilitates efficient intra-modal and inter-modal interactions. Specifically, we design a novel Hierarchical Interaction Mamba (\textbf{HIM}) to efficiently capture the hierarchical characteristics of WSIs and transcriptomic data. HIM extracts coarse-grained features through the aggregation of fine-grained instances and enables efficient bidirectional interactions for multi-grained instances. In this way, SurvMamba can extract enhanced local information from numerous fine-grained instances and global information from coarse-grained instances, resulting in more comprehensive intra-modal information to predict patient survival outcomes. Furthermore, we propose an Interaction Fusion Mamba (\textbf{IFM}) to facilitate interactions between histological and genomic features across different granularities, thereby providing refined intra-modal representations at both fine and coarse levels.  Finally, these multi-grained features are adaptively integrated to formulate the final survival prediction. The main contributions of this paper can be summarized as follows:

\begin{itemize}
\item This work is the first attempt to introduce the Mamba model into multi-modal survival prediction, effectively processing high-dimensional WSIs and transcriptomic data with promising performance.
\item We propose a Hierarchical Interaction Mamba module to efficiently encode more comprehensive intra-modal representations at both fine-grained and coarse-grained levels from WSI and transcriptomic data.
\item We introduce an Interaction Fusion Mamba module, designed to facilitate interaction and integration of histological and genomic features across various levels, thereby capturing multi-modal features from diverse perspectives.
\item Extensive experiments are conducted on five public TCGA datasets, and results show that SurvMamba outperforms a variety of state-of-the-art methods with a smaller computational cost. 
\end{itemize}

\section{Related Work}

\subsection{Multi-modal Survival Prediction}
Survival outcome prediction, also known as time-to-event analysis~\cite{david2012survival}, concentrates on probabilistic assessment of experiencing a specified event such as mortality in the clinical setting before a time under both uncensored and right-censored data. Right-censored data represents cases wherein the event of interest remains unobserved throughout the duration of the study. In the current state-of-the-art methods, estimating cancer patient survival heavily depends on physicians' assessment of histology and/or interpretation of genomic sequencing report~\cite{chen2021multimodal}. Consequently, there is a growing interest in multi-modal learning methods that integrate histopathology and genomic data for survival prediction~\cite{chen2020pathomic, chen2021whole, chen2021multimodal, subramanian2021multimodal, chen2022pan, jaume2023modeling, xu2023multimodal}. Chen et al.~\cite{chen2021multimodal} proposed a Multi-modal Co-Attention Transformer framework that identifies informative instances from pathological images using genomic features as queries. Qiu et al.~\cite{qiu2023deep}
proposed PONET, a novel pathology-genomic deep model informed by biological pathway that integrates pathological images and genomic data to improve survival prediction. Jaume et al.~\cite{jaume2023modeling} modeled interactions between biological pathway and histology patch tokens using a memory-efficient multi-modal Transformer for survival analysis. Zhang et al.~\cite{zhang2024prototypical} proposed a new framework named Prototypical Information Bottlenecking and Disentangling (PIBD), including Prototypical Information Bottleneck module for intra-modal redundancy and Prototypical Information Disentanglement module for inter-modal redundancy. Existing methods mainly regard multi-modal survival prediction as a fine-grained recognition task, focusing on histological patches and genomic functions to model fine-grained cross-modal relationships with local information. Neglecting coarser-grained information may result in overlooking some essential global features vital for accurately predicting patient survival. WSIs and genomic data exhibit an intrinsic hierarchical structure, where different granularity levels yield unique and critical insights into patient prognosis. 


\subsection{State Space Models}
Recently, the State Space Models (SSMs) have shown significant effectiveness of state space transformation in capturing the dynamics and dependencies of language sequences. The structured state-space sequence model (S4) introduced in \cite{gu2021efficiently} is specifically designed to model long-range dependencies, offering the advantage of linear complexity. 
Various models including S5~\cite{smith2022simplified}, H3~\cite{fu2022hungry} and GSS~\cite{mehta2022long} have been developed base on S4, and Mamba~\cite{gu2023mamba} distinguishes itself by introducing a data-dependent SSM layer and a selection mechanism using parallel scan (S6). Compared to Transformers with quadratic-complexity attention, Mamba excels at processing long sequences with linear complexity. Existing models based on MIL for multi-modal survival prediction struggle with enabling effective and efficient interactions among vast numbers of instances~\cite{ilse2018attention, li2021dual, shao2021transmil, li2024generalizable}, burdened by significant computational requirements. The introduction of Mamba addresses these challenges by incorporating input-adaptive and global information modeling techniques that emulate self-attention functionalities while retaining linear complexity. This breakthrough diminishes the computational overhead and provides a potential multi-modal learning framework for survival prediction.

\section{Methodology}
\begin{figure*}[t]
  \centering
  \includegraphics[width=\linewidth]{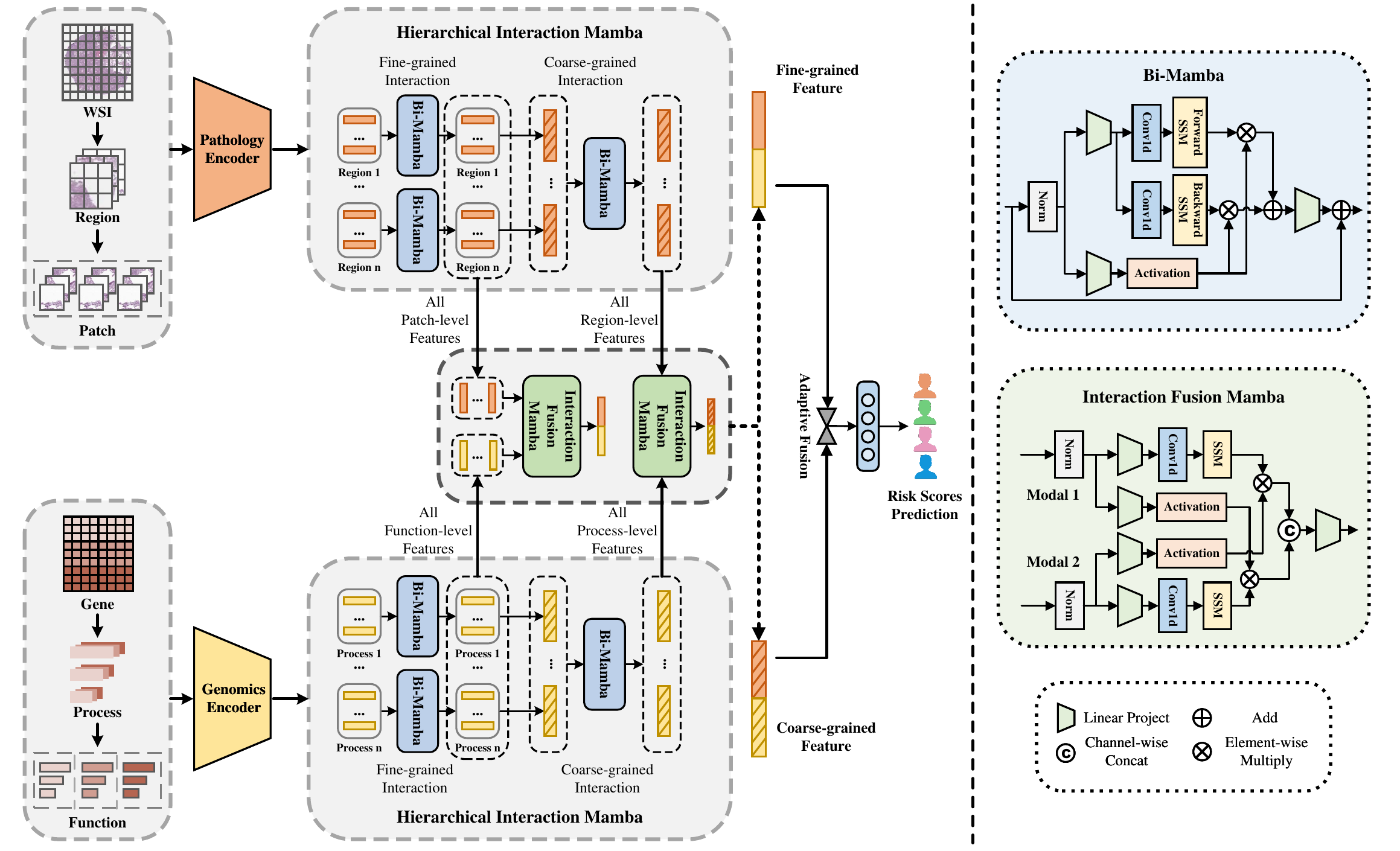}
  \caption{Overview of SurvMamba architecture. WSIs and transcriptomics are illustrated in a three-layer structure, comprising WSI/Region/Patch for WSIs and Gene/Process/Function for transcriptomics, respectively. }
  \label{fig:overview}
\end{figure*}

In this section, we first describe the preliminaries of the state space model (SSM) and then provide an overview of our proposed SurvMamba (shown in Figure~\ref{fig:overview}) and its core components.

\subsection{Preliminaries}

The SSM-based models, \textit{i.e.}, structured state-space sequence models (S4), and Mamba, have emerged as promising architectures for modeling long sequences with linear complexity. With four parameters $(\mathbf{\Delta}, \mathbf{A}, \mathbf{B}, \mathbf{C})$, they map an input stimulus $x(t) \in \mathbb{R}$ to an output response $y(t) \in \mathbb{R}$ through an intermediate latent state $h(t) \in \mathbb{R}^N$. This process can be illustrated in the following equations:

\begin{equation}
\begin{split}
h'(t) &= \mathbf{A}h(t) + \mathbf{B}x(t) \\
y(t) &= \mathbf{C}h(t)
\end{split}
\label{eq1}
\end{equation}

where $\mathbf{A} \in \mathbb{R}^{N \times N}$ denotes the evolution parameter, and $\mathbf{B} \in \mathbb{R}^{N}$, $\mathbf{C} \in \mathbb{R}^{N}$ represent projection parameters. The models exploit a timescale parameter $\mathbf{\Delta}$ to convert continuous parameters $\mathbf{A}$, $\mathbf{B}$ into their discrete counterparts $\bar{\mathbf{A}}$, $\bar{\mathbf{B}}$, according to:

\begin{equation}
\begin{split}
\mathbf{\bar{A}} &= \exp(\mathbf{\Delta} \mathbf{A}) \\
\mathbf{\bar{B}} &= (\mathbf{\Delta} \mathbf{A})^{-1}(\exp(\mathbf{\Delta} \mathbf{A}) - I) \cdot \mathbf{\Delta} \mathbf{B}.
\end{split}
\end{equation}

Subsequently, these discrete parameters enable the reformulation of Eq.~(\ref{eq1}) in a recurrent format, facilitating efficient autoregressive inference:
\begin{equation}
\begin{split}
h_t &= \bar{\mathbf{A}}h_{t-1} +\bar{\mathbf{B}}x_t, \\
y_t &= \mathbf{C}h_t.
\end{split}
\end{equation}
Finally, the output is derived through global convolution. $M$ is the length of the input sequence $x$ and $\mathbf{\bar{K}} \in \mathbb{R}^{M}$ is a structured convolutional kernel:
\begin{equation}
\begin{split}
\mathbf{\bar{K}} &= (\mathbf{C}\mathbf{\bar{B}}, \mathbf{C}\bar{\mathbf{AB}}, \ldots, \mathbf{C}\bar{\mathbf{A}}^{M-1}\mathbf{\bar{B}}), \\
\mathbf{y} &= \mathbf{x} * \mathbf{\bar{K}}.
\end{split}
\end{equation}

\subsection{Overview and Problem Formulation}

As illustrated in Figure~\ref{fig:overview}, we propose a novel state space model with multi-grained multi-modal interaction named \textbf{SurvMamba} for survival prediction. Clinical data for each patient is encapsulated in a quadruple $X_i = (I_i, G_i, c_i, t_i)$, where $I_i$ represent the set of WSIs, $G_i$ refers to the set of transcriptomics, $c_i \in \{0, 1\}$ is the censoring status and $t_i \in \mathbb{R}^+$ denotes overall survival time (in months). Our objective is to employ WSI $I_i$ and transcriptomic data $G_i$ to estimate hazard functions $f^i_{hazard}(t)$, which represent the probability of a death event occurring in a brief interval following time $t$ for the $i$-th patient.

To capture more comprehensive multi-modal representations, we exploit the inherent hierarchical structure of WSI and transcriptomic data to extract multi-grained information, subsequently enabling efficient interaction and fusion among them through Mamba-based modules. We adopt the MIL framework to learn pathological and genomic representations, formulating each WSI and transcriptomics as a ``bag''. We extract features of fine-grained instances (Patch or Function) from the Pathology Encoder and the Genomics Encoder in groups. Then, with bidirectional Mamba in dual-level MIL framework, the Hierarchical Interaction Mamba (\textbf{HIM}) module aggregates fine-grained instances into coarse-grained instances (Region or Process) and enables efficient intra-modal interactions at different granularities. Further, the Interaction Fusion Mamba (\textbf{IFM}) module facilitates multi-grained interactions between histological and genomic features to derive fine-grained and coarse-grained inter-modal representations. Finally, multi-grained multi-modal features will be adaptively fused to predict a hazard function and get the survival risk scores.

\subsection{Hierarchical Multi-modal Representations}
\label{sec:HMR}

WSIs and transcriptomics exhibit hierarchical structure with multi-level prognostic insights. In this study, we harness this hierarchical nature by representing WSIs and transcriptomics through a three-layer structure. Specifically, WSI is formulated with structures of WSI-, region-, and patch-level, while transcriptomic data are structured across gene-, function-, and process-level, respectively. 

For an original WSI $I$, we split it into $M$ non-overlapping regions $R$ with size of $4,096 \times 4,096$, and each region is further split into $N$ non-overlapping patches $P$ with size of $256 \times 256$, where $I = \{R_1, R_2, \ldots, R_M\}$, and $R_m = \{P_{m1}, P_{m2}, \ldots, P_{mN}\}, 1 \leq m \leq M$. We extract the patch-level features with a frozen pre-trained Pathology Encoder $f_I(\cdot)$, resulting in patch-level token sequence $T^I_m$ in $R_m$, $T^I_m=\{f_I(P_{m1}), f_I(P_{m2}), \ldots, f_I(P_{mN})\}$.

Given a set of transcriptomics measurements $G$, it can be mapped into $J$ different biological processes $S$. The $j$-th process $S_j$ contains $K_j$ genomic functions $F$, where $G = \{S_1, S_2, \ldots, S_J\}$, and $S_j = \{F_{j1}, F_{j2}, \ldots, F_{jK_j}\}, 1 \leq j \leq J$. The number of genomic functions contained within each process varies. We encode genomic functions with Genomics Encoder $f_G(\cdot)$ which contains multilayer perceptrons (MLPs) with learnable weights, resulting in function-level token sequence $T^G_j$ in $S_j$, $T^G_j=\{f_G(F_{j1}), f_G(F_{j2}), \ldots, f_G(F_{jK_j})\}$.

\subsection{Hierarchical Interaction Mamba}
\label{sec:HIM}

Structural WSI and transcriptomic data exhibit certain dependencies within local or global features. To capture these dependencies, 
we develop the HIM module. Inspired by~\cite{zhu2024vision}, this module is designed to learn intra-model features by integrating a bidirectional Mamba (Bi-Mamba) mechanism within a dual-level MIL framework. The specific structure of Bi-Mamba is shown in the top-right corner of Figure~\ref{fig:overview}. Our approach aims to model correlations across features of varying granularities effectively. For the first level MIL, the HIM module regards fine-grained (\textit{i.e.}, patch- and function-level) features $T^I_i$ and $T^G_j$ as instances, using Bi-Mamba with shared parameters between groups (\textit{i.e.}, region or process) to model fine-grained histological and genomic long sequences, resulting in enhanced fine-grained features ${T^I_m}^{\prime}$ and ${T^G_j}^{\prime}$: 
\begin{equation}
\begin{split}
{T^I_m}^{\prime} &= \textbf{Bi-Mamba}(T^I_m) \\
{T^G_j}^{\prime} &= \textbf{Bi-Mamba}(T^G_j).
\end{split}
\end{equation}
Subsequently, with the second level MIL, features ${T^I_m}^{\prime}$ and ${T^G_j}^{\prime}$ are aggregated into coarse-grained (\emph{i.e.}, region- and process-level) features by average pooling, and Bi-Mamba is then re-applied to facilitate interactions among them to get improved coarse-grained features $T^I$ and $T^G$. Through this dual-level deep interaction mechanism for each modality, the HIM ensures a thorough and efficient integration of local fine-grained and global coarse-grained features, obtaining enhanced and comprehensive intra-modal histological and genomic features.

\begin{equation}
\begin{split}
T^I &= \textbf{Bi-Mamba}([\text{Pool}({T^I_1}^{\prime}), \text{Pool}({T^I_2}^{\prime}), \ldots, 	\text{Pool}({T^I_M}^{\prime})]) \\
T^G &= \textbf{Bi-Mamba}([\text{Pool}({T^G_1}^{\prime}), \text{Pool}({T^G_2}^{\prime}), \ldots, \text{Pool}({T^G_J}^{\prime})]).
\end{split}
\end{equation}


\subsection{Interaction Fusion Mamba}
\label{sec:IFM}

To facilitate cross-modal feature interaction and fusion at different granularities, we introduce an IFM module, which is shown in the middle-right part of Figure~\ref{fig:overview}. In IFM, we project features from two modalities and employ gating mechanisms to encourage complementary feature learning from each other while suppressing redundant features. After that, multi-modal features are concentrated to form a fused representation. Cross-modality communications are established at fine-grained and coarse-grained level via cascaded IFM. Through this block, we can obtain fine-grained fused features $H_f$ from features ${T^I}^{\prime}$ and ${T^G}^{\prime}$, and get coarse-grained fused features $H_c$ from features $T^I$ and $T^G$, as follows:
\begin{equation}
\begin{split}
H_f &= \textbf{IFM}([{T^I}^{\prime}, {T^G}^{\prime}]) \\
H_c &= \textbf{IFM}([T^I, T^G]).
\end{split}
\end{equation}

\subsection{Survival Prediction}
\label{sec:Prediction}

Recognizing the varying significance of features at different granularities for survival prediction, we employ an adaptive fusion strategy with learnable weight denoted as $\alpha$ to fuse features across granular levels. We initialize $\alpha$ at 0.5 and include it in the optimizer parameters, adjusting it during optimization to minimize the loss function. Subsequently, we derive the final feature set for prognostication with $H_f$ and $H_c$, as follows:
\begin{equation}
    H = \alpha H_f + (1-\alpha) H_c.
\end{equation}
Survival prediction estimates the risk probability of an outcome event before a specific time. For the final multi-modal feature $H^{i}$ of \( i \)-th patient, we use NLL loss ~\cite{zadeh2020bias} as the loss function for optimizing survival prediction, following~\cite{xu2023multimodal}.

\section{Experiments}


\subsection{Datasets and Settings}
\noindent \textbf{Datasets.} 
To demonstrate the performance of our proposed method, we conducted a series of experiments using five public cancer datasets from The Cancer Genome Atlas (TCGA)\footnote{\url{http://www.cancer.gov/tcga}}, which include paired diagnostic WSIs and transcriptomic data alongside verified survival outcomes. The datasets encompass Breast Invasive Carcinoma (BRCA) with 869 cases, Bladder Urothelial Carcinoma (BLCA) with 359 cases, Colon and Rectum Adenocarcinoma (COADREAD) with 296 cases, Uterine Corpus Endometrial Carcinoma (UCEC) with 480 cases, and Lung Adenocarcinoma (LUAD) with 453 cases. Regarding transcriptomic data, we identify 352 unique genomic functions and 42 biological processes, as cataloged in the Kyoto Encyclopedia of Genes and Genomes database (KEGG)\footnote{\url{https://www.genome.jp/kegg/}}. 
\\
\noindent \textbf{Evaluation Metrics.} 
The models are evaluated using the concordance index (c-index), where a higher value indicates better performance. This index quantifies the proportion of all possible pairs of observations for which the model accurately predicts the sequence of actual survival outcomes.
\\
\noindent \textbf{Implementation.}
For WSI preprocessing, we segmented each WSI into regions of $4096\times4096$ pixels at a magnification level of $20\times$, subsequently subdividing these regions into patches of $256\times256$ pixels. The RAdam optimizer was utilized to facilitate model optimization, with a batch size set to 1, a learning rate of $2 \times 10^{-4}$, and a weight decay parameter of $5\times10^{-3}$. The Pathology Encoder~\cite{wang2021transpath} generates 768-dimensional embeddings that are projected to 512 dimensions. The Genomics Encoder comprises a two-layer feed-forward network designed to produce genomic tokens with 512 dimensions. We set SSM dimension to 16. To enhance the robustness of model training, 5-fold cross-validation was applied on all models. All computational experiments were conducted on an NVIDIA-A800 GPU.

\subsection{Comparisons with the State-of-the-Art}
\noindent \textbf{Baselines.} 
To perform a comprehensive comparison with our method, we implemented and evaluated some latest survival prediction methods, including: (1) Unimodal baselines. For transcriptomic data, we specifically implemented SNN~\cite{klambauer2017self} and SNNTrans~\cite{klambauer2017self, kleinbaum1996survival}. For histology, we compared the SOTA MIL methods ABMIL~\cite{ilse2018attention}, CLAM~\cite{lu2021data}, TransMIL~\cite{shao2021transmil}, R\textsuperscript{2}T-MIL~\cite{tang2024feature}. (2) Multi-modal baselines. We compared SOTA methods for multi-modal survival outcome prediction with the previous set-based network architectures (CLAM, TransMIL) with concatenation (Cat) and Kronecker product (KP), two common late fusion mechanisms to integrate bag-level WSI features and genomic features as multi-modal baselines. We also compared CMTA~\cite{zhou2023cross}, MCAT~\cite{chen2021multimodal}, MOTCat~\cite{xu2023multimodal} and SurvPath~\cite{jaume2023modeling} four SOTA methods for multi-modal survival outcome prediction. Table~\ref{tab:sota} shows the experimental results of all methods on all five TCGA datasets. 


\begin{table*}[h]
 \centering
  \caption{c-index (mean ± std) over five TCGA datasets. G. and H. refer to genomic modality (transcriptomics) and histological modality (WSI), respectively. The best results and the second-best results are highlighted in \textbf{bold} and in \underline{underline}.}
  \label{tab:tesult_comparison}
  \resizebox{\linewidth}{!}{
  \begin{tabular}{c|cc|cccccc}
    \hline
    Model & G. & H. & BRCA & BLCA   & COADREAD & UCEC & LUAD & Overall\\
    \hline
    SNN  &  \checkmark  & & 0.606$\pm$0.011 & 0.610$\pm$0.038 & 0.617$\pm$0.025 & 0.610$\pm$0.032 & 0.589$\pm$0.031 & 0.606\\
    SNNTrans &  \checkmark  &  & 0.621$\pm$0.032 & 0.611$\pm$0.010 & 0.635$\pm$0.044 & 0.592$\pm$0.017 & 0.602$\pm$0.044 & 0.612\\
    \midrule
    ABMIL &  & \checkmark &  0.613$\pm$0.033 & 0.588$\pm$0.033 & 0.624$\pm$0.050 & 0.618$\pm$0.014 & 0.604$\pm$0.043 & 0.609\\
    CLAM-SB &  & \checkmark & 0.605$\pm$0.062 & 0.602$\pm$0.031 & 0.598$\pm$0.036 & 0.576$\pm$0.043 & 0.586$\pm$0.033 & 0.593\\
    CLAM-MB &  & \checkmark & 0.611$\pm$0.041 & 0.609$\pm$0.010 & 0.611$\pm$0.036 & 0.589$\pm$0.023 & 0.612$\pm$0.022 & 0.606\\
    TransMIL &  & \checkmark & 0.628$\pm$0.015 & 0.604$\pm$0.045 & 0.627$\pm$0.425 & 0.601$\pm$0.030 & 0.626$\pm$0.030 & 0.617\\
    {R\textsuperscript{2}T-MIL} &  & \checkmark & 0.641$\pm$0.006 & 0.611$\pm$0.041 & 0.604$\pm$0.029 & 0.618$\pm$0.030 & 0.624$\pm$0.029 & 0.620\\ 
    \midrule
    CLAM-MB (Cat) & \checkmark & \checkmark & 0.628$\pm$0.047 & 0.619$\pm$0.032 & 0.614$\pm$0.021 & 0.601$\pm$0.031 & 0.610$\pm$0.032 & 0.614\\
    CLAM-MB (KP) & \checkmark & \checkmark & 0.655$\pm$0.045 & 0.633$\pm$0.027 & 0.651$\pm$0.053 & 0.637$\pm$0.021 & 0.629$\pm$0.061 & 0.641\\
    TransMIL (Cat) & \checkmark & \checkmark & 0.651$\pm$0.039 & 0.631$\pm$0.031 & 0.636$\pm$0.026 & 0.622$\pm$0.043 & 0.641$\pm$0.033 & 0.636\\
    TransMIL (KP)  & \checkmark & \checkmark & 0.671$\pm$0.021 & 0.656$\pm$0.038 & 0.661$\pm$0.034 & 0.649$\pm$0.036 & 0.652$\pm$0.054 & 0.658\\
    CMTA & \checkmark & \checkmark & 0.687$\pm$0.015 & 0.689$\pm$0.035 & 0.663$\pm$0.040 & 0.685$\pm$0.006 & 0.672$\pm$0.025 & 0.679\\
    MCAT & \checkmark & \checkmark & 0.670$\pm$0.032 & 0.669$\pm$0.026 & 0.667$\pm$0.025 & 0.660$\pm$0.032 & 0.682$\pm$0.042 & 0.670\\
    MOTCat & \checkmark & \checkmark & 0.692$\pm$0.036 & 0.688$\pm$0.029 & 0.669$\pm$0.042 & 0.692$\pm$0.024 & \underline{0.687$\pm$0.046} & 0.686\\
    SurvPath & \checkmark & \checkmark & \underline{0.713$\pm$0.025} & \underline{0.707$\pm$0.014} & \underline{0.683$\pm$0.022} & \underline{0.720$\pm$0.026} & 0.684$\pm$0.025 & \underline{0.701}\\ \hline
    \textbf{SurvMamba} & \checkmark & \checkmark & \textbf{0.737$\pm$0.014} & \textbf{0.720$\pm$0.027} & \textbf{0.697$\pm$0.018} & \textbf{0.731$\pm$0.012} & \textbf{0.702$\pm$0.020} & \textbf{0.717}\\ 
  \hline
\end{tabular}
}
\label{tab:sota}
\end{table*}

\noindent \textbf{Unimodal v.s. Multi-modal.} 
Compared with all unimodal methods, our proposed SurvMamba achieved the highest performance in all five datasets, indicating effective integration of multi-modal features in our method. In comparison with methods for genomic data, SNN and SNNTrans, SurvMamba outperformed them on all benchmarks, with overall c-index performance increases of 11.1\% and 10.5\%, respectively. Against the pathology baselines, SurvMamba improved on all the pathology-based unimodal approaches, with performance improvements in the overall c-index ranging 10.0\% to 12.4\%, demonstrating the merit of integrating histopathology and genomic features. The comparison results also highlight the benefits of utilizing multi-modality in survival prediction.

\noindent \textbf{Multi-modal SOTA v.s. SurvMamba.} SurvMamba outperforms all multi-modal approaches with an overall c-index performance increase ranging from 1.6\% to 10.3\%. In comparative analyses within each dataset, SurvMamba achieved the highest c-index performance in four out of five cancer benchmarks, demonstrating its potential as a general method for survival prediction task. When compared with MIL-based multi-modal methods with different fusion methods, SurvMamba achieved a performance increase in overall c-index ranging from 5.9\% to 10.3\%, highlighting the effectiveness of the proposed multi-modal integration method. Besides, our model also  outperforms other SOTA multi-modal learning methods by a obvious margin, including CMAT, MCAT, MOTCat, and SurvPath, showcasing its outstanding ability for multi-modal learning for prognosis.

\begin{figure}[ht]
  \centering
  \includegraphics[width=0.7\linewidth]{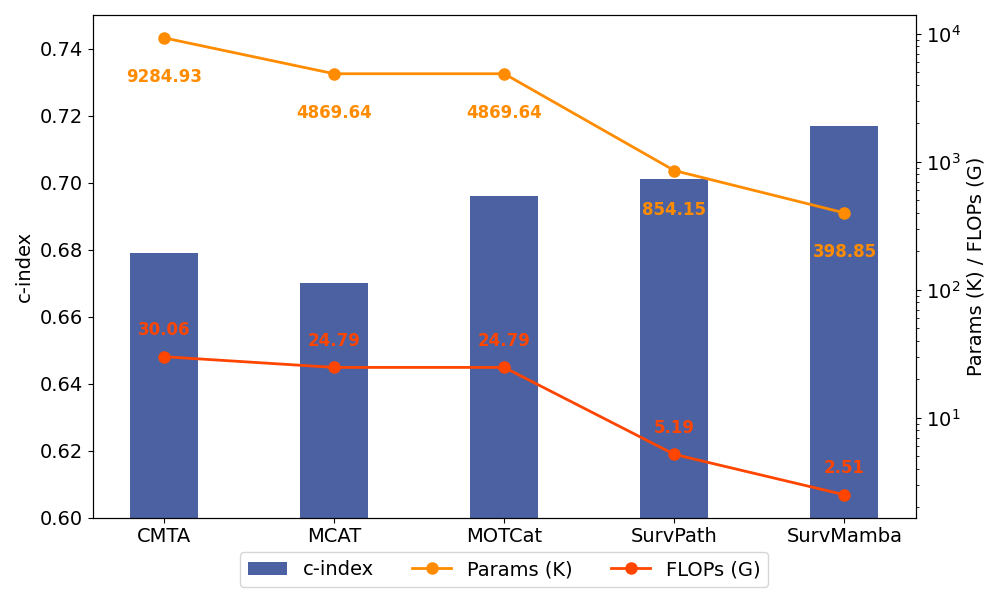}
  \caption{Computational complexity analysis. }
  \label{fig:complexity}
\end{figure}

\noindent \textbf{Computational Cost Comparison.} 
Existing frameworks for multi-modal survival prediction are based on the attention mechanism to model the intra- and inter-relationships, while our method leverages the state space model. To validate the computational efficiency of our method, we compared SurvMamba with some multi-modal SOTA methods which show the top 5 on the overall c-index of the five TCGA datasets. The computational complexity is evaluated using model parameters (Params) and floating-point operation count (FLOPs). Params evaluates the network’s scale, while FLOPs assess the model’s complexity. As shown in Figure~\ref{fig:complexity}, we report the overall c-index on the TCGA dataset, Params, and FLOPs for different SOTA models. Compared to CMTA, the model size and GPU memory of SurvMamba are reduced by 95.70\% and 91.65\% with a 3.8\% better overall c-index. Compared to the second best method SurvPath, SurvMamba also decreases by 53.3\% and 51.6\% in model size and GPU memory with a 1.6\% better c-index. These results show that our model achieves superior performance compared to the state-of-the-art methods with reduced computational cost.

\subsection{Ablation Study}

\noindent \textbf{Ablation of components.}
We conducted ablation studies to validate the effectiveness of the proposed modules. Detailed experimental setups are as follows:

\begin{table*}[h]
  \centering
  \caption{We compare the effects of different components on the performance (c-index) of SurvMamba.}
  \resizebox{\linewidth}{!}{
  \begin{tabular}{ccccccccccc}
    \toprule
    \multirow{2}{*}{\begin{tabular}[c]{@{}c@{}}Model\end{tabular} }  &\multicolumn{2}{c}{HIM} & \multicolumn{2}{c}{IFM} & \multirow{2}{*}{\begin{tabular}[c]{@{}c@{}}BRCA\end{tabular} }  & \multirow{2}{*}{\begin{tabular}[c]{@{}c@{}}BLCA\end{tabular} }   & \multirow{2}{*}{\begin{tabular}[c]{@{}c@{}}COADREAD\end{tabular} } & \multirow{2}{*}{\begin{tabular}[c]{@{}c@{}}UCEC\end{tabular} } & \multirow{2}{*}{\begin{tabular}[c]{@{}c@{}}LUAD\end{tabular} } & \multirow{2}{*}{\begin{tabular}[c]{@{}c@{}}Overall\end{tabular} }\\
    \cmidrule(r){2-3} \cmidrule(r){4-5}
    & fine & coarse & fine & coarse \\
    \midrule
     A & & & & & 0.688$\pm$0.023 & 0.667$\pm$0.031 & 0.660$\pm$0.031 & 0.686$\pm$0.015 & 0.668$\pm$0.047 & 0.674 \\
     B & \checkmark & & & & 0.708$\pm$0.011 & 0.683$\pm$0.065 & 0.666$\pm$0.022 & 0.717$\pm$0.023 & 0.674$\pm$0.309 & 0.689\\
     C & \checkmark & \checkmark & & & 0.707$\pm$0.007 & 0.691$\pm$0.030 & 0.681$\pm$0.030 & 0.719$\pm$0.020 & 0.675$\pm$0.033 & 0.695\\
     D & \checkmark & \checkmark & \checkmark & & 0.711$\pm$0.081 & 0.700$\pm$0.034 & 0.684$\pm$0.036 & 0.721$\pm$0.017 & 0.689$\pm$0.016 & 0.701\\
     E & \checkmark & \checkmark & & \checkmark & 0.709$\pm$0.010 & 0.696$\pm$0.013 & 0.687$\pm$0.030 & 0.720$\pm$0.010 & 0.679$\pm$0.019 & 0.698\\
     F & \checkmark & \checkmark & \checkmark & \checkmark & 0.737$\pm$0.014 & 0.720$\pm$0.027 & 0.697$\pm$0.018 & 0.731$\pm$0.012 & 0.702$\pm$0.020 & 0.717\\
    \bottomrule
  \end{tabular}
  }
  \label{tab:ab_study}
\end{table*}

\begin{enumerate}
    \item[(A)] \textbf{None}: All fine-grained histological or genomic features are fed as an input vector into the Bi-Mamba block. These unimodal features are then concatenated to predict survival outcomes. 
    \item[(B)] \textbf{Fine-grained HIM}: Utilizes shared Bi-Mamba in HIM to refine fine-grained features across specific groups, enhancing these features for prediction without considering coarse-grained information.
    \item[(C)] \textbf{Multi-grained HIM}: Extends Model (B) by incorporating coarse-grained information and integrating multi-grained data for prediction.
    \item[(D)] \textbf{Multi-grained HIM + Fine-grained IFM}: Model (C) with IFM for fine-grained features.    
    \item[(E)] \textbf{Multi-grained HIM + Coarse-grained IFM}: Model (C) with IFM for coarse-grained features.
    \item[(F)] \textbf{Multi-grained HIM + Multi-grained IFM}: Model (C) with IFM for multi-grained features (\emph{i.e.}, SurvMamba).
\end{enumerate}

Table~\ref{tab:ab_study} illustrates that Model B, by grouping and processing fine-grained features rather than directly learning from extensive sequences of these features like Model A, more effectively identifies nuanced local relationships. This methodological shift results in a c-index increase from 0.674 to 0.689. The introduction of coarse-grained information, representing broader characteristics in Model C further enhances the c-index to 0.695. These outcomes highlight the advantage of a hierarchical approach in harnessing multi-grained information for prognostic purposes. Improving upon Model C, the application of IFM to integrate and interact with fine- or coarse-grained features resulted in a c-index of 0.701 and 0.689, respectively. This demonstrates the IFM module's effectiveness in integrating multi-modal features. Our proposed SurvMamba model, facilitating both intra-modal and inter-modal interactions and integrations across different granularity levels, delivers a notable c-index of 0.717. Results of the ablation study illustrate the benefits of integrating fine and coarse-grained features and the promising SSM architecture for survival outcome prediction.

\noindent \textbf{Ablation of Mamba.}
We conducted an ablation study by replacing Mamba modules with Transformer to demonstrate the effectiveness of using the Mamba model for WSIs and transcriptomic data. We measured both the inference runtime and FLOPs for the Transformer-based and Mamba-based approaches. As shown in Table~\ref{tab_add_ab}, the Mamba-based method demonstrated greater performance with reduced computational costs.

\begin{table*}[h]
  \centering
    \caption{We ablate Mamba modules by replacing them with Transformer models.}  
  \resizebox{\linewidth}{!}{
  \begin{tabular}{cccccccc}
    \toprule
    Model & BRCA & BLCA & COADREAD & UCEC & LUAD & Time & Flops\\
    \midrule
    {Transformer-based} & 0.721$\pm$0.016 & 0.702$\pm$0.046 & 0.696$\pm$0.021 & 0.705$\pm$0.052 & 0.679$\pm$0.035 & 0.78s & 9.73G\\ 
    {Mamba-based}  & 0.737$\pm$0.014 & 0.720$\pm$0.027 & 0.697$\pm$0.018 & 0.731$\pm$0.012 & 0.702$\pm$0.020 & 0.22s & 2.51G\\
    \bottomrule
  \end{tabular}
  }
  \label{tab_add_ab}
\end{table*}

\begin{figure}[h]
  \centering
  \includegraphics[width=0.6\linewidth]{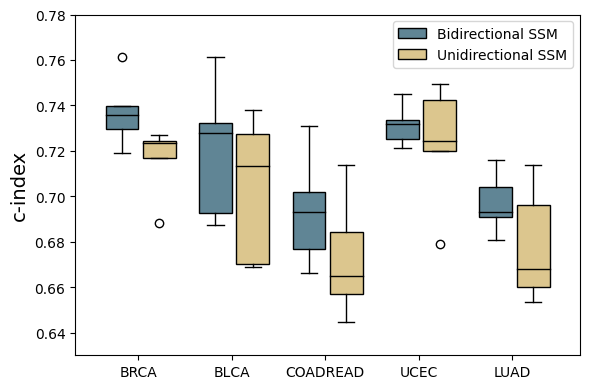}
  \caption{Ablation study on the bidirectional SSM designed in SurvMamba.}
  \label{fig:bi-SSM}
\end{figure}

\noindent \textbf{Impact of Bidirectional and Unidirectional SSM.}
Figure~\ref{fig:bi-SSM} demonstrates that integrating bidirectionality into state space models significantly enhances prognostic predictions, as indicated by a higher median c-index and a narrower interquartile range. This effect is particularly pronounced in the BRCA dataset, where the boxplots show minimal overlap, suggesting a robust improvement. This improvement is likely due to the model's enhanced ability to capture complex, bidirectional dependencies within histological and genomic features.

\subsection{Survival Analysis}
\begin{figure*}[ht]
  \centering
  \subfloat[BRCA]
  {
      \label{fig:BRCA_risk}
      \includegraphics[width=0.18\textwidth]{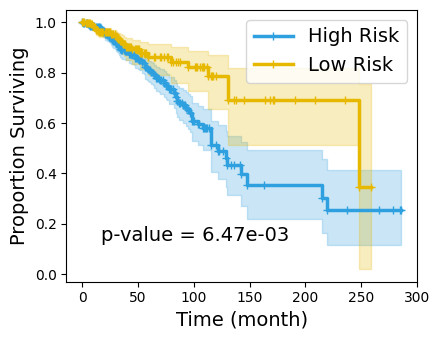}
  }
  \subfloat[BLCA]
  {
      \label{fig:BLCA_risk}
      \includegraphics[width=0.18\textwidth]{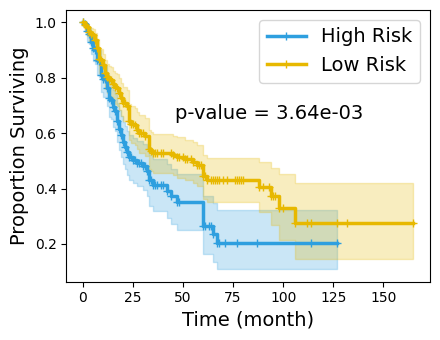}
  }
  \subfloat[COADREAD]
  {
      \label{fig:COADREAD}
      \includegraphics[width=0.18\textwidth]{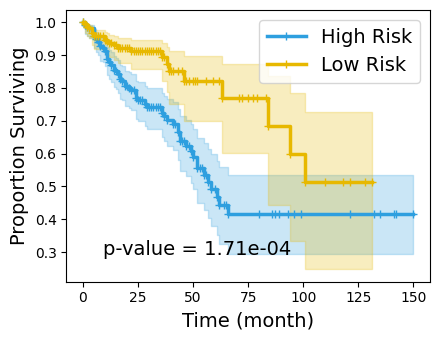}
  }
  \subfloat[UCEC]
  {
      \label{fig:UCEC}
      \includegraphics[width=0.18\textwidth]{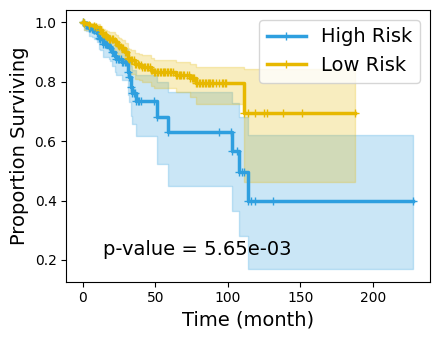}
  }
  \subfloat[LUAD]
  {
      \label{fig:LUAD}
      \includegraphics[width=0.18\textwidth]{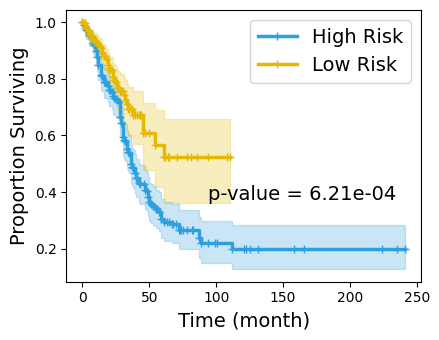}
  }
  \caption{Kaplan-Meier survival curves of SurvMamba on five TCGA cancer datasets.}  
  \label{fig:KM}            
\end{figure*}

To further validate the effectiveness of SurvMamba for survival analysis, we divided all patients into a low-risk group and a high-risk group based on the median value of the predicted risk scores generated by SurvMamba. Then, we utilize Kaplan-Meier analysis to visualize the survival events of all patients. Meanwhile, we also employ the Logrank test ($p$-value) to measure the statistical significance between the low-risk group and the high-risk group. Shared areas within two groups refer to the confidence intervals, and a $p$-value of less than 0.05 indicates significant statistical difference. As shown in Figure~\ref{fig:KM}, patients in the low-risk (yellow) and high-risk (blue) groups are stratified clearly on all datasets, demonstrating the prognostic value of SurvMamba in predicting patient outcomes and guiding treatment decisions.  

\section{Conclusion}
In this paper, we proposed a novel state space model with multi-grained multi-modal interaction, termed SurvMamba, for survival prediction from WSIs and transcriptomic data. 
SurvMamba utilizes multi-grained information from hierarchical structures in WSIs and transcriptomic data. We introduced a HIM module that facilitates efficient interactions between intra-modal features at different granularity levels, thereby enriching the unimodal representations. Furthermore, we introduced an IFM module to integrate inter-modal features across various levels, capturing more comprehensive multi-modal features for survival analysis. The experimental results demonstrate SurvMamba's superiority in both performance and computational efficiency, underlining its potential for clinical utilization, such as informing diagnostic and treatment choices for cancer patients.

\bibliographystyle{abbrvnat}
\bibliography{main}

\begin{thebibliography}{41}
\providecommand{\natexlab}[1]{#1}
\providecommand{\url}[1]{\texttt{#1}}
\expandafter\ifx\csname urlstyle\endcsname\relax
  \providecommand{\doi}[1]{doi: #1}\else
  \providecommand{\doi}{doi: \begingroup \urlstyle{rm}\Url}\fi

\bibitem[Chen et~al.(2020)Chen, Lu, Wang, Williamson, Rodig, Lindeman, and Mahmood]{chen2020pathomic}
R.~J. Chen, M.~Y. Lu, J.~Wang, D.~F. Williamson, S.~J. Rodig, N.~I. Lindeman, and F.~Mahmood.
\newblock Pathomic fusion: an integrated framework for fusing histopathology and genomic features for cancer diagnosis and prognosis.
\newblock \emph{IEEE Transactions on Medical Imaging}, 41\penalty0 (4):\penalty0 757--770, 2020.

\bibitem[Chen et~al.(2021{\natexlab{a}})Chen, Lu, Shaban, Chen, Chen, Williamson, and Mahmood]{chen2021whole}
R.~J. Chen, M.~Y. Lu, M.~Shaban, C.~Chen, T.~Y. Chen, D.~F. Williamson, and F.~Mahmood.
\newblock Whole slide images are 2d point clouds: Context-aware survival prediction using patch-based graph convolutional networks.
\newblock In \emph{Medical Image Computing and Computer Assisted Intervention--MICCAI 2021: 24th International Conference, Strasbourg, France, September 27--October 1, 2021, Proceedings, Part VIII 24}, pages 339--349. Springer, 2021{\natexlab{a}}.

\bibitem[Chen et~al.(2021{\natexlab{b}})Chen, Lu, Weng, Chen, Williamson, Manz, Shady, and Mahmood]{chen2021multimodal}
R.~J. Chen, M.~Y. Lu, W.-H. Weng, T.~Y. Chen, D.~F. Williamson, T.~Manz, M.~Shady, and F.~Mahmood.
\newblock Multimodal co-attention transformer for survival prediction in gigapixel whole slide images.
\newblock In \emph{Proceedings of the IEEE/CVF International Conference on Computer Vision}, pages 4015--4025, 2021{\natexlab{b}}.

\bibitem[Chen et~al.(2022{\natexlab{a}})Chen, Chen, Li, Chen, Trister, Krishnan, and Mahmood]{chen2022scaling}
R.~J. Chen, C.~Chen, Y.~Li, T.~Y. Chen, A.~D. Trister, R.~G. Krishnan, and F.~Mahmood.
\newblock Scaling vision transformers to gigapixel images via hierarchical self-supervised learning.
\newblock In \emph{Proceedings of the IEEE/CVF Conference on Computer Vision and Pattern Recognition}, pages 16144--16155, 2022{\natexlab{a}}.

\bibitem[Chen et~al.(2022{\natexlab{b}})Chen, Lu, Williamson, Chen, Lipkova, Noor, Shaban, Shady, Williams, Joo, et~al.]{chen2022pan}
R.~J. Chen, M.~Y. Lu, D.~F. Williamson, T.~Y. Chen, J.~Lipkova, Z.~Noor, M.~Shaban, M.~Shady, M.~Williams, B.~Joo, et~al.
\newblock Pan-cancer integrative histology-genomic analysis via multimodal deep learning.
\newblock \emph{Cancer Cell}, 40\penalty0 (8):\penalty0 865--878, 2022{\natexlab{b}}.

\bibitem[David and Mitchel(2012)]{david2012survival}
G.~K. David and K.~Mitchel.
\newblock Survival analysis: a self-learning text, 2012.

\bibitem[Fu et~al.(2022)Fu, Dao, Saab, Thomas, Rudra, and R{\'e}]{fu2022hungry}
D.~Y. Fu, T.~Dao, K.~K. Saab, A.~W. Thomas, A.~Rudra, and C.~R{\'e}.
\newblock Hungry hungry hippos: Towards language modeling with state space models.
\newblock \emph{arXiv preprint arXiv:2212.14052}, 2022.

\bibitem[Fu et~al.(2020)Fu, Jung, Torne, Gonzalez, V{\"o}hringer, Shmatko, Yates, Jimenez-Linan, Moore, and Gerstung]{fu2020pan}
Y.~Fu, A.~W. Jung, R.~V. Torne, S.~Gonzalez, H.~V{\"o}hringer, A.~Shmatko, L.~R. Yates, M.~Jimenez-Linan, L.~Moore, and M.~Gerstung.
\newblock Pan-cancer computational histopathology reveals mutations, tumor composition and prognosis.
\newblock \emph{Nature cancer}, 1\penalty0 (8):\penalty0 800--810, 2020.

\bibitem[Gu and Dao(2023)]{gu2023mamba}
A.~Gu and T.~Dao.
\newblock Mamba: Linear-time sequence modeling with selective state spaces.
\newblock \emph{arXiv preprint arXiv:2312.00752}, 2023.

\bibitem[Gu et~al.(2021)Gu, Goel, and R{\'e}]{gu2021efficiently}
A.~Gu, K.~Goel, and C.~R{\'e}.
\newblock Efficiently modeling long sequences with structured state spaces.
\newblock \emph{arXiv preprint arXiv:2111.00396}, 2021.

\bibitem[Gui et~al.(2023)Gui, Chen, Zhao, Cao, Liu, Xiong, Yu, Liao, Cao, Li, et~al.]{gui2023multimodal}
C.-P. Gui, Y.-H. Chen, H.-W. Zhao, J.-Z. Cao, T.-J. Liu, S.-W. Xiong, Y.-F. Yu, B.~Liao, Y.~Cao, J.-Y. Li, et~al.
\newblock Multimodal recurrence scoring system for prediction of clear cell renal cell carcinoma outcome: a discovery and validation study.
\newblock \emph{The Lancet Digital Health}, 5\penalty0 (8):\penalty0 e515--e524, 2023.

\bibitem[Ietswaart et~al.(2021)Ietswaart, Gyori, Bachman, Sorger, and Churchman]{ietswaart2021genewalk}
R.~Ietswaart, B.~M. Gyori, J.~A. Bachman, P.~K. Sorger, and L.~S. Churchman.
\newblock Genewalk identifies relevant gene functions for a biological context using network representation learning.
\newblock \emph{Genome biology}, 22:\penalty0 1--35, 2021.

\bibitem[Ilse et~al.(2018)Ilse, Tomczak, and Welling]{ilse2018attention}
M.~Ilse, J.~Tomczak, and M.~Welling.
\newblock Attention-based deep multiple instance learning.
\newblock In \emph{International conference on machine learning}, pages 2127--2136. PMLR, 2018.

\bibitem[Jaume et~al.(2023)Jaume, Vaidya, Chen, Williamson, Liang, and Mahmood]{jaume2023modeling}
G.~Jaume, A.~Vaidya, R.~Chen, D.~Williamson, P.~Liang, and F.~Mahmood.
\newblock Modeling dense multimodal interactions between biological pathways and histology for survival prediction.
\newblock \emph{arXiv preprint arXiv:2304.06819}, 2023.

\bibitem[Kanehisa and Goto(2000)]{kanehisa2000kegg}
M.~Kanehisa and S.~Goto.
\newblock Kegg: kyoto encyclopedia of genes and genomes.
\newblock \emph{Nucleic acids research}, 28\penalty0 (1):\penalty0 27--30, 2000.

\bibitem[Ke et~al.(2023)Ke, Shen, Lu, Guo, and Shen]{ke2023mine}
J.~Ke, Y.~Shen, Y.~Lu, Y.~Guo, and D.~Shen.
\newblock Mine local homogeneous representation by interaction information clustering with unsupervised learning in histopathology images.
\newblock \emph{Computer Methods and Programs in Biomedicine}, 235:\penalty0 107520, 2023.

\bibitem[Klambauer et~al.(2017)Klambauer, Unterthiner, Mayr, and Hochreiter]{klambauer2017self}
G.~Klambauer, T.~Unterthiner, A.~Mayr, and S.~Hochreiter.
\newblock Self-normalizing neural networks.
\newblock \emph{Advances in neural information processing systems}, 30, 2017.

\bibitem[Kleinbaum and Klein(1996)]{kleinbaum1996survival}
D.~G. Kleinbaum and M.~Klein.
\newblock \emph{Survival analysis a self-learning text}.
\newblock Springer, 1996.

\bibitem[Li et~al.(2021)Li, Li, and Eliceiri]{li2021dual}
B.~Li, Y.~Li, and K.~W. Eliceiri.
\newblock Dual-stream multiple instance learning network for whole slide image classification with self-supervised contrastive learning.
\newblock In \emph{Proceedings of the IEEE/CVF conference on computer vision and pattern recognition}, pages 14318--14328, 2021.

\bibitem[Li et~al.(2024)Li, Chen, Chen, Yu, Yang, Wang, Ding, and Han]{li2024generalizable}
H.~Li, Y.~Chen, Y.~Chen, R.~Yu, W.~Yang, L.~Wang, B.~Ding, and Y.~Han.
\newblock Generalizable whole slide image classification with fine-grained visual-semantic interaction.
\newblock In \emph{Proceedings of the IEEE/CVF Conference on Computer Vision and Pattern Recognition}, pages 11398--11407, 2024.

\bibitem[Li et~al.(2022)Li, Wu, Li, and Wang]{li2022hfbsurv}
R.~Li, X.~Wu, A.~Li, and M.~Wang.
\newblock Hfbsurv: hierarchical multimodal fusion with factorized bilinear models for cancer survival prediction.
\newblock \emph{Bioinformatics}, 38\penalty0 (9):\penalty0 2587--2594, 2022.

\bibitem[Lu et~al.(2021)Lu, Williamson, Chen, Chen, Barbieri, and Mahmood]{lu2021data}
M.~Y. Lu, D.~F. Williamson, T.~Y. Chen, R.~J. Chen, M.~Barbieri, and F.~Mahmood.
\newblock Data-efficient and weakly supervised computational pathology on whole-slide images.
\newblock \emph{Nature biomedical engineering}, 5\penalty0 (6):\penalty0 555--570, 2021.

\bibitem[Mehta et~al.(2022)Mehta, Gupta, Cutkosky, and Neyshabur]{mehta2022long}
H.~Mehta, A.~Gupta, A.~Cutkosky, and B.~Neyshabur.
\newblock Long range language modeling via gated state spaces.
\newblock \emph{arXiv preprint arXiv:2206.13947}, 2022.

\bibitem[Milanez-Almeida et~al.(2020)Milanez-Almeida, Martins, Germain, and Tsang]{milanez2020cancer}
P.~Milanez-Almeida, A.~J. Martins, R.~N. Germain, and J.~S. Tsang.
\newblock Cancer prognosis with shallow tumor rna sequencing.
\newblock \emph{Nature medicine}, 26\penalty0 (2):\penalty0 188--192, 2020.

\bibitem[Qiu et~al.(2023)Qiu, Khormali, and Liu]{qiu2023deep}
L.~Qiu, A.~Khormali, and K.~Liu.
\newblock Deep biological pathway informed pathology-genomic multimodal survival prediction.
\newblock \emph{arXiv preprint arXiv:2301.02383}, 2023.

\bibitem[Shao et~al.(2019)Shao, Han, Cheng, Cheng, Wang, Sun, Lu, Zhang, Zhang, and Huang]{shao2019integrative}
W.~Shao, Z.~Han, J.~Cheng, L.~Cheng, T.~Wang, L.~Sun, Z.~Lu, J.~Zhang, D.~Zhang, and K.~Huang.
\newblock Integrative analysis of pathological images and multi-dimensional genomic data for early-stage cancer prognosis.
\newblock \emph{IEEE transactions on medical imaging}, 39\penalty0 (1):\penalty0 99--110, 2019.

\bibitem[Shao et~al.(2021)Shao, Bian, Chen, Wang, Zhang, Ji, et~al.]{shao2021transmil}
Z.~Shao, H.~Bian, Y.~Chen, Y.~Wang, J.~Zhang, X.~Ji, et~al.
\newblock Transmil: Transformer based correlated multiple instance learning for whole slide image classification.
\newblock \emph{Advances in neural information processing systems}, 34:\penalty0 2136--2147, 2021.

\bibitem[Smith et~al.(2022)Smith, Warrington, and Linderman]{smith2022simplified}
J.~T. Smith, A.~Warrington, and S.~W. Linderman.
\newblock Simplified state space layers for sequence modeling.
\newblock \emph{arXiv preprint arXiv:2208.04933}, 2022.

\bibitem[Smug et~al.(2023)Smug, Szczepaniak, Rocha, Dunin-Horkawicz, and Mostowy]{smug2023ongoing}
B.~J. Smug, K.~Szczepaniak, E.~P. Rocha, S.~Dunin-Horkawicz, and R.~J. Mostowy.
\newblock Ongoing shuffling of protein fragments diversifies core viral functions linked to interactions with bacterial hosts.
\newblock \emph{Nature Communications}, 14\penalty0 (1):\penalty0 7460, 2023.

\bibitem[Subramanian et~al.(2021)Subramanian, Syeda-Mahmood, and Do]{subramanian2021multimodal}
V.~Subramanian, T.~Syeda-Mahmood, and M.~N. Do.
\newblock Multimodal fusion using sparse cca for breast cancer survival prediction.
\newblock In \emph{2021 IEEE 18th International Symposium on Biomedical Imaging (ISBI)}, pages 1429--1432. IEEE, 2021.

\bibitem[Tang et~al.(2024)Tang, Zhou, Huang, Zhu, Zhang, and Liu]{tang2024feature}
W.~Tang, F.~Zhou, S.~Huang, X.~Zhu, Y.~Zhang, and B.~Liu.
\newblock Feature re-embedding: Towards foundation model-level performance in computational pathology.
\newblock In \emph{Proceedings of the IEEE/CVF Conference on Computer Vision and Pattern Recognition}, pages 11343--11352, 2024.

\bibitem[Tong et~al.(2023)Tong, Lin, Yang, Song, Zhang, Xie, Tian, Luo, Liang, Huang, et~al.]{tong2023prioritizing}
M.~Tong, Y.~Lin, W.~Yang, J.~Song, Z.~Zhang, J.~Xie, J.~Tian, S.~Luo, C.~Liang, J.~Huang, et~al.
\newblock Prioritizing prognostic-associated subpopulations and individualized recurrence risk signatures from single-cell transcriptomes of colorectal cancer.
\newblock \emph{Briefings in Bioinformatics}, 24\penalty0 (3):\penalty0 bbad078, 2023.

\bibitem[Tran et~al.(2021)Tran, Kondrashova, Bradley, Williams, Pearson, and Waddell]{tran2021deep}
K.~A. Tran, O.~Kondrashova, A.~Bradley, E.~D. Williams, J.~V. Pearson, and N.~Waddell.
\newblock Deep learning in cancer diagnosis, prognosis and treatment selection.
\newblock \emph{Genome Medicine}, 13:\penalty0 1--17, 2021.

\bibitem[Wang et~al.(2021)Wang, Yang, Zhang, Wang, Zhang, Huang, Yang, and Han]{wang2021transpath}
X.~Wang, S.~Yang, J.~Zhang, M.~Wang, J.~Zhang, J.~Huang, W.~Yang, and X.~Han.
\newblock Transpath: Transformer-based self-supervised learning for histopathological image classification.
\newblock In \emph{Medical Image Computing and Computer Assisted Intervention--MICCAI 2021: 24th International Conference, Strasbourg, France, September 27--October 1, 2021, Proceedings, Part VIII 24}, pages 186--195. Springer, 2021.

\bibitem[Xu and Chen(2023)]{xu2023multimodal}
Y.~Xu and H.~Chen.
\newblock Multimodal optimal transport-based co-attention transformer with global structure consistency for survival prediction.
\newblock In \emph{Proceedings of the IEEE/CVF International Conference on Computer Vision}, pages 21241--21251, 2023.

\bibitem[Zadeh and Schmid(2020)]{zadeh2020bias}
S.~G. Zadeh and M.~Schmid.
\newblock Bias in cross-entropy-based training of deep survival networks.
\newblock \emph{IEEE transactions on pattern analysis and machine intelligence}, 43\penalty0 (9):\penalty0 3126--3137, 2020.

\bibitem[Zhang et~al.(2024{\natexlab{a}})Zhang, Schroeder, Yan, Yang, Hu, Lee, Cho, Susztak, Xu, Feldman, et~al.]{zhang2024inferring}
D.~Zhang, A.~Schroeder, H.~Yan, H.~Yang, J.~Hu, M.~Y. Lee, K.~S. Cho, K.~Susztak, G.~X. Xu, M.~D. Feldman, et~al.
\newblock Inferring super-resolution tissue architecture by integrating spatial transcriptomics with histology.
\newblock \emph{Nature Biotechnology}, pages 1--6, 2024{\natexlab{a}}.

\bibitem[Zhang et~al.(2021)Zhang, Hu, and Smith]{zhang2021protocol}
X.~Zhang, Y.~Hu, and D.~R. Smith.
\newblock Protocol for hsdfinder: Identifying, annotating, categorizing, and visualizing duplicated genes in eukaryotic genomes.
\newblock \emph{STAR protocols}, 2\penalty0 (3):\penalty0 100619, 2021.

\bibitem[Zhang et~al.(2024{\natexlab{b}})Zhang, Xu, Chen, Xie, and Chen]{zhang2024prototypical}
Y.~Zhang, Y.~Xu, J.~Chen, F.~Xie, and H.~Chen.
\newblock Prototypical information bottlenecking and disentangling for multimodal cancer survival prediction.
\newblock \emph{arXiv preprint arXiv:2401.01646}, 2024{\natexlab{b}}.

\bibitem[Zhou and Chen(2023)]{zhou2023cross}
F.~Zhou and H.~Chen.
\newblock Cross-modal translation and alignment for survival analysis.
\newblock In \emph{Proceedings of the IEEE/CVF International Conference on Computer Vision}, pages 21485--21494, 2023.

\bibitem[Zhu et~al.(2024)Zhu, Liao, Zhang, Wang, Liu, and Wang]{zhu2024vision}
L.~Zhu, B.~Liao, Q.~Zhang, X.~Wang, W.~Liu, and X.~Wang.
\newblock Vision mamba: Efficient visual representation learning with bidirectional state space model.
\newblock \emph{arXiv preprint arXiv:2401.09417}, 2024.

\end{thebibliography}

\end{document}